\documentclass[sigconf]{acmart}    

\pdfoutput=1

\usepackage{multirow, booktabs}
\usepackage{hyperref}   
\usepackage{cleveref}   
\usepackage{mathrsfs}
\usepackage{dsfont}
\usepackage{arydshln} 
\usepackage{arydshln} 
\usepackage{balance}
\usepackage{hyperref}
\AtBeginDocument{%
  }

\copyrightyear{2025} 
\acmYear{2025} 
\setcopyright{acmlicensed}\acmConference[MM '25]{Proceedings of the 33rd
ACM International Conference on Multimedia}{October 27--31, 2025}{Dublin,
Ireland}
\acmBooktitle{Proceedings of the 33rd ACM International Conference on
Multimedia (MM '25), October 27--31, 2025, Dublin, Ireland}
\acmDOI{10.1145/3746027.3755324}
\acmISBN{979-8-4007-2035-2/2025/10}

\begin{document}

\title{Mitigating Information Loss under High Pruning Rates for Efficient Large Vision Language Models}


\author{Mingyu Fu}
\affiliation{
  \institution{Northwestern Polytechnical University}    
  \city{Xi'an}
  \state{Shaanxi}
  \country{China}
}
\affiliation{
  \institution{National Engineering Laboratory for Integrated Aero-Space-Ground-Ocean Big Data Application Technology}
  \city{Xi'an}
  \state{Shaanxi}
  \country{China}
}
\email{fumy@mail.nwpu.edu.cn}
\authornote{Both authors contributed equally to this research.}

\author{Wei Suo}
\affiliation{
  \institution{ Northwestern Polytechnical University}   
  \city{Xi'an}
  \state{Shaanxi}
  \country{China}
}
\affiliation{
  \institution{National Engineering Laboratory for Integrated Aero-Space-Ground-Ocean Big Data Application Technology}
  \city{Xi'an}
  \state{Shaanxi}
  \country{China}
}
\email{suowei1994@mail.nwpu.edu.cn}
\authornotemark[1]
\authornote{Corresponding author}

\author{Ji Ma}
\affiliation{
  \institution{ Northwestern Polytechnical University}     
  \city{Xi'an}
  \state{Shaanxi}
  \country{China}
}
\affiliation{
  \institution{National Engineering Laboratory for Integrated Aero-Space-Ground-Ocean Big Data Application Technology}
  \city{Xi'an}
  \state{Shaanxi}
  \country{China}
}
\email{maji@mail.nwpu.edu.cn}

\author{Lin Yuanbo Wu}
\affiliation{
  \institution{ Swansea University}    
  \city{Swansea}
  \state{Wales}
  \country{UK}
}
\email{xiaoxian.wu9188@gmail.com}

\author{Peng Wang}
\affiliation{
  \institution{ Northwestern Polytechnical University}   
  \city{Xi'an}
  \state{Shaanxi}
  \country{China}
}
\affiliation{
  \institution{National Engineering Laboratory for Integrated Aero-Space-Ground-Ocean Big Data Application Technology}
  \city{Xi'an}
  \state{Shaanxi}
  \country{China}
}
\email{peng.wang@nwpu.edu.cn}

\author{Yanning Zhang}
\affiliation{
  \institution{ Northwestern Polytechnical University}   
  \city{Xi'an}
  \state{Shaanxi}
  \country{China}
}
\affiliation{
  \institution{National Engineering Laboratory for Integrated Aero-Space-Ground-Ocean Big Data Application Technology}
  \city{Xi'an}
  \state{Shaanxi}
  \country{China}
}
\email{ynzhang@nwpu.edu.cn}

\renewcommand{\shortauthors}{Mingyu Fu et al.}

\begin{abstract}

Despite the great success of Large Vision Language Models (LVLMs), their high computational cost severely limits their broad applications. The computational cost of LVLMs mainly stems from the visual sequence of the input, which consists of hundreds or even thousands of tokens. 
Although existing methods have made progress by removing redundant tokens, they suffer from severe performance degradation with high pruning rates due to the loss of visual information.  
In this paper, we propose an \textbf{A}daptive \textbf{C}ontent \textbf{C}ompensation \textbf{M}ethod (ACCM),  which can effectively mitigate the visual information loss via an image caption.
Specifically, ACCM comprises two key components: a lightweight caption model and a selector. Firstly the caption model generates question-related descriptions under the guidance of the user instruction. Then the selector further identifies a contextually appropriate caption from multiple candidates.  
Leveraging self-supervised learning, our modules could be learned efficiently without any human or automated labeling. 
We conduct extensive experiments across seven benchmarks and the results show that ACCM significantly outperforms existing methods with lower FLOPs (\textit{e.g.}, surpassing SOTA by 20.6\% with 6.5\% fewer FLOPs) \footnote{\url{https://github.com/ASGO-MM/ACCM}}.

\end{abstract}





\maketitle

\section{Introduction}
Large Vision Language Models (LVLMs) have rapidly evolved as a transformative force in artificial intelligence in recent years \cite{liu2023visual,li2023blip,zhu2023minigpt,alayrac2022flamingo}. Built upon Large Language Models (LLMs) \cite{radford2018improving,brown2020language,zhang2022opt,touvron2023llama}, LVLMs integrate text and images to achieve cross-modal understanding and generation, obtaining remarkable improvements across a range of tasks (\textit{e.g.}, visual reasoning and visual grounding). Their influence is becoming increasingly profound, with applications spanning healthcare, education and autonomous systems \cite{team2023gemini, achiam2023gpt, lu2024deepseek}. While LVLMs have demonstrated remarkable capabilities across multiple domains, the huge computational overheads seriously hinder their practical applications. 
In contemporary LVLMs, the vision sequence typically contains hundreds to thousands of tokens, especially when processing high-resolution images \cite{ye2024mplug,zhang2024beyond,meng2024deepstack}. Furthermore, in the Transformer architecture \cite{vaswani2017attention}, the computational cost grows quadratically with the length of the input sequence. As a result, the heavy computational burden of LVLMs  primarily originates from lengthy visual sequences.

\begin{figure}
  \centering
    \includegraphics[width=0.5\textwidth]{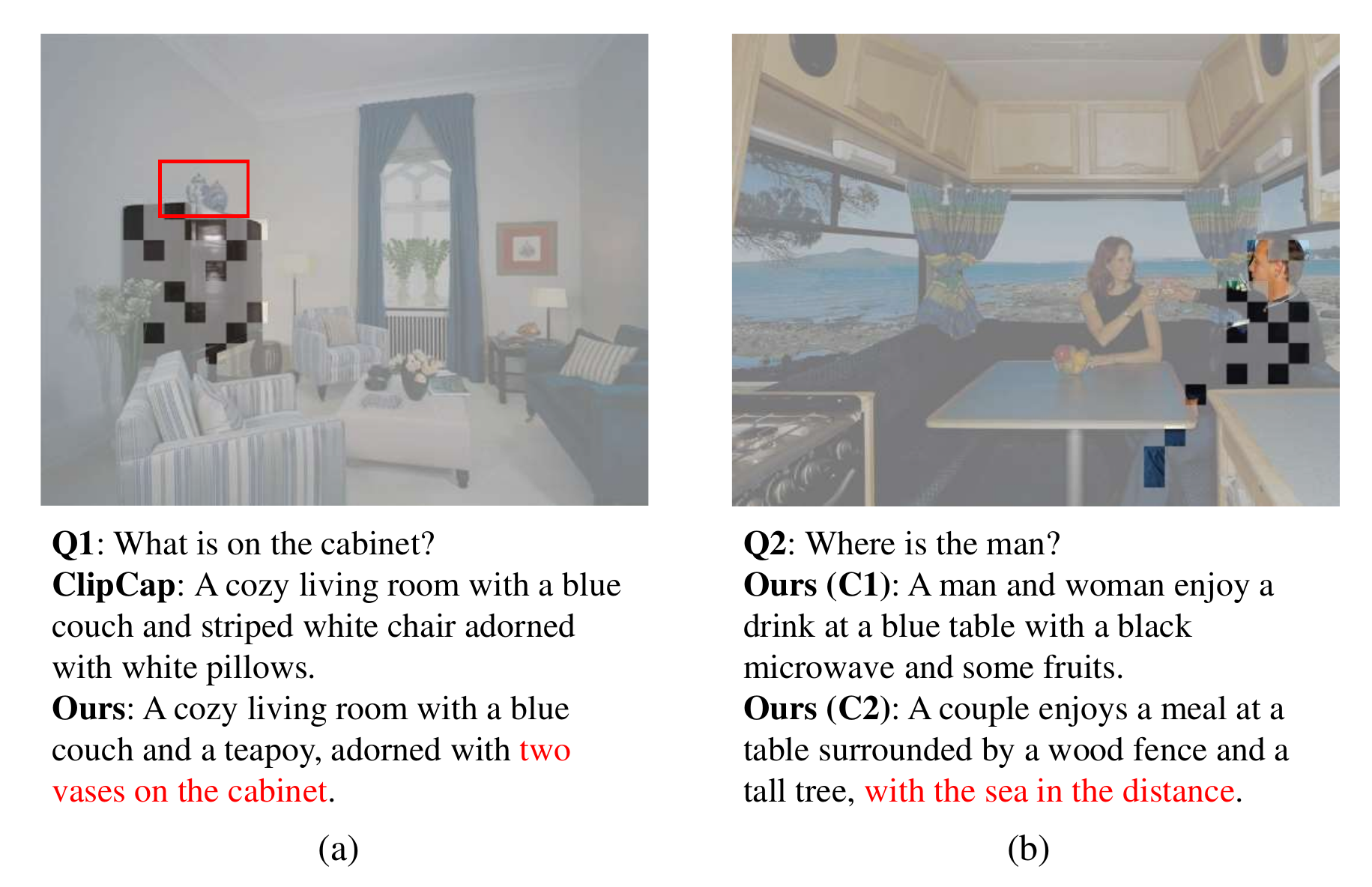}
   
  \caption{We apply FastV with 97\% pruning rate on LLaVA-1.5-7B to demonstrate the pruned results. Key information for answering questions is highlighted in color. The red box indicates the key area for answering Q1. (a) Our generated caption correctly focuses on the cabinet area to answer the question Q1. (b) Both captions (C1 and C2) are relevant to the question Q2, while C2 focuses on the outdoors and is more proper for answering Q2.}

  \label{fig:pic-1}
\end{figure}

In order to mitigate the computational burden of LVLMs, a number of token pruning approaches \cite{chen2024image,lin2024boosting,chen2024efficient,shang2024llava,li2024tokenpacker,qian2024spatial,liu2024multi,zhang2025llava,suo2024pruning} have been proposed recently. They prove the high redundancy of the visual sequence in LVLMs and employ various strategies to compress it. Many of these approaches \cite{chen2024image,shang2024llava,liu2024multi,huang2024dynamic,xing2024pyramiddrop} focus on selecting the most critical tokens based on specific criteria. Others  \cite{li2024tokenpacker,qian2024spatial,zhang2025llava} compress visual sequences by designing efficient modules (\textit{e.g.}, multi-modal connector).
Although contemporary approaches significantly reduce visual token redundancy, they share a common problem: the performance degrades seriouly at high pruning rates. For instance, FastV \cite{chen2024image} exhibits a performance drop of over 30\% when the pruning rate exceeds 90\% (a detailed discussion can be found in \cref{performance degradation}).

In fact, the above results are intuitive. At high pruning rates, most visual tokens will be discarded, while the remaining tokens retain insufficient visual information to support multi-modal perception and reasoning. As illustrated in Figure \ref{fig:pic-1} (a), at 97\% pruning rate, limited visual tokens cannot cover the entire cabinet area, inevitably resulting in the loss of other visual information. As a result, the pruned LVLMs fail to correctly answer the question (Q1), which requires the model to pay attention to the two vases on top of the cabinet.

To mitigate the visual information loss, a straightforward idea is to use an image caption to recover the lost information. However, a general caption is sub-optimal for supplementing the lost information (detailed results are demonstrated in \cref{ablation}). To investigate the underlying reasons for this phenomenon, we conduct a systematic qualitative analysis and obtain the following insights: 
Firstly, for the same image, different questions typically require various visual information. A general image caption may not inherently contain the information required to answer the question. As illustrated in Figure \ref{fig:pic-1} (a), the first caption (generated by ClipCap \cite{mokady2021clipcap}) primarily describes the sofa area, whereas the second caption (Ours) focuses on the cabinet region, which is directly relevant to the question (Q1). Therefore, how to generate question-related captions is a key problem. 
Secondly, while generated captions remain relevant to the question, they exhibit diverse expressive forms with varying semantic emphases.
Among these variants, certain captions probably provide more appropriate  content than others.  
As illustrated in Figure \ref{fig:pic-1} (b), the first caption (C1) focuses on interior furnishings (\textit{i.e.}, table and microwave) and the second caption (C2) emphasizes on outdoor environments (\textit{i.e.}, tree and sea). Both captions describe the man's surroundings, while the latter is more proper for answering the question (Q2).
Thus, how to identify a contextually appropriate caption is crucial to supplement the lost visual information.

Based on the aforementioned analysis, we propose an \textbf{A}daptive \textbf{C}ontent \textbf{C}ompensation \textbf{M}ethod (ACCM), which can effectively mitigate the visual information loss under high pruning rates. In contrast to other methods, ACCM recovers the lost information adaptively by an image caption. Specifically, ACCM consists of two modules, a lightweight image caption model and a selector. Firstly the caption model generates question-related descriptions based on discarded visual tokens under the guidance of the corresponding question. 
Then the selector further identifies the most contextually appropriate caption from multiple candidates.
Finally the caption model and selector are jointly optimized using Direct Preference Optimization (DPO) \cite{rafailov2023direct} due to the non-differentiable nature of the selection operation.
More importantly, both modules are trained in a self-supervised manner, without the need for extensive manual or automated labeling.

We conduct extensive experiments across various multi-modal tasks, including visual question answering, hallucination evaluation and image caption. Experimental results show that our method effectively supplements the lost visual information under high pruning rates, outperforming other methods significantly with lower FLOPs. For instance, ACCM surpasses PyramidDrop \cite{xing2024pyramiddrop} by 20.6\% with 6.5\% fewer FLOPs.

Our main contributions can be summarized as follows:

{\bf (1)} To mitigate the severe visual information loss under high pruning rates, we propose to adaptively supplement the lost information leveraging an image caption.  Benefiting from self-supervised learning, our approach efficiently recovers the lost visual information without any human or automated labeling.

{\bf (2)} Our approach, ACCM, introduces two newly designed modules: a lightweight caption model and a selector. The caption model generates question-related descriptions under the guidance of the user instruction. The selector further identifies a contextually appropriate caption from multiple candidates.

{\bf (3)} We perform comprehensive evaluations on multiple benchmarks. Experimental results show that our approach significantly surpasses existing methods with lower FLOPs (\textit{e.g.}, outperforming the state-of-the-art method by 20.6\% with 6.5\% fewer FLOPs).

\section{Related Work}
\subsection{Large Vision Language Models}
 
\begin{figure*}[ht]
  \centering
    \includegraphics[width=\linewidth]{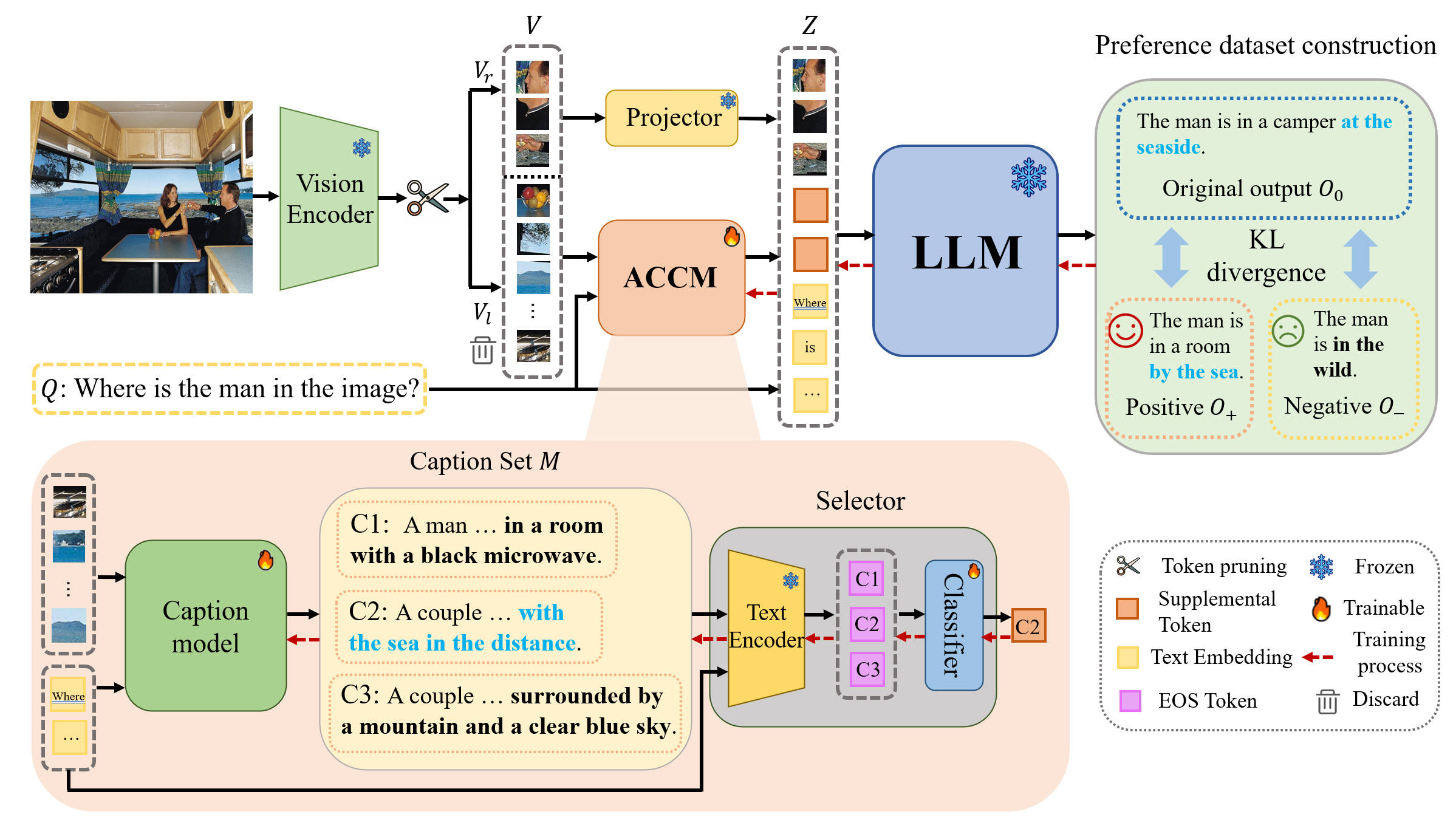}     
   
  \caption{Overview of our approach. Our ACCM consists of two components: a lightweight caption model and a selector. To be specific, firstly the caption model converts discarded visual tokens $V_l$ into question-related captions under the guidance of  the user instruction $Q$. Then the selector further chooses a contextually appropriate description from caption set $M$. Finally, we construct the preference dataset by calculating KL divergence between original output $O_0$ and output of pruned sequence $Z$ and apply DPO to optimize our model. Question-related information in captions is bolded and key information for answering the question is highlighted in color.}
 
  \label{fig:pic-2}
\end{figure*}

Recent advances in Large Language Models (LLMs) have enabled the development of Large Vision Language Models (LVLMs) capable of cross-modal reasoning with images, videos, and audio. Early models like CLIP \cite{radford2021learning} aligned vision and language, while later systems such as Flamingo \cite{alayrac2022flamingo}, BLIP-2 \cite{li2023blip}, LLaVA \cite{li2023blip}, and MiniGPT-4 \cite{zhu2023minigpt} introduced mechanisms (e.g., Q-Formers, linear projections) to bridge modality gaps more efficiently. However, high computational costs remain a challenge due to the large number of visual tokens and the quadratic growth of self-attention overhead with respect to input length. Moreover, several studies \cite{chen2024image,zhang2025llava} have demonstrated the presence of substantial redundancy within visual sequences. Thus, efficient token compression is essential to ensure the scalability of LVLMs under limited resources.

\subsection{Token Pruning Methods}

Token pruning is a key technique for improving transformer efficiency by adaptively removing redundant tokens based on contextual relevance. Widely studied in natural language processing (NLP) and computer vision (CV), it has recently gained traction in LVLMs. In NLP, methods like Funnel-Transformer \cite{dai2020funnel} and Pyramid-BERT \cite{huang2022pyramid} progressively condense sequences, while PoWER-BERT \cite{goyal2020power} prunes tokens based on similarity. In CV, ToMe \cite{bolya2022token} merges similar tokens via soft matching, PPT \cite{wu2023ppt} unifies pruning and merging within a whole framework, and Zero-TPrune \cite{wang2024zero} uses PageRank algorithm for token pruning.

In the area of LVLMs, token pruning methods fall into two categories: methods without modifying LVLMs weights and requiring re-training. In the former, FastV \cite{chen2024image} removes half of the visual tokens after the second LLM layer using average attention scores. VTW \cite{lin2024boosting} discards tokens at a specific layer based on KL divergence. LLaVolta \cite{chen2024efficient} compresses visual tokens with average pooling and multi-stage training. In the latter, LLaVA-PruMerge \cite{shang2024llava} identifies key tokens via the Interquartile Range (IQR) method \cite{boukerche2020outlier} and merges them using key clustering. TokenPacker \cite{li2024tokenpacker} introduces a region-to-point attention module for compressing multi-scale features. However, these methods often suffer from performance drops at high pruning rates due to information loss. In contrast, our approach adaptively compensates for missing visual information using an image caption generated by a question-guided caption model and selector.

\section{Method}
In this section, we first review the architecture of LVLMs in \cref{sec:background}. Then we present our ACCM in \cref{sec:xxx}. Finally, we introduce the optimization process of our model in \cref{sec:optimization}.
\subsection{Background}
\label{sec:background}

The prevailing architecture of contemporary LVLMs typically comprises three fundamental components: a vision encoder, a multi-modal connector, and a large language model (LLM). Given an image $I$, the vision encoder (\textit{e.g.}, CLIP ViT \cite{radford2021learning}) first encodes $I$ into a visual sequence $V$. 
Then the multi-modal connector (\textit{e.g.}, MLP in LLaVA-1.5 \cite{liu2024improved}) projects $V$ into the same semantic space with the LLM. 
Finally the $V$ is concatenated with user instruction $Q$ and sent to the LLM for multi-modal comprehension and generation. For the sake of simplicity, the systematic prompt tokens are omitted in this work.
The generation process could be formulated as:
\begin{equation}
  p\left(Y \mid V, Q\right)=\prod_{i=1}^L p_{\theta}\left(y_i \mid V, Q, Y, _{<i}\right),
\end{equation}
where $\theta$ represents parameters of the LLM and $Y$ is the answer with $L$ tokens. The visual sequence $V$ typically comprises hundreds to thousands of tokens, much longer than $Q$. Moreover, in the Transformer architecture \cite{vaswani2017attention}, the computational cost grows quadratically with the length of the input sequence. 
As a result, the visual sequence $V$ severely compromises the computational efficiency of LVLMs. 

To enhance the computational efficiency of LVLMs, several recent works \cite{chen2024image,lin2024boosting,chen2024efficient,shang2024llava,li2024tokenpacker} demonstrate the high redundancy of the visual sequence and employ various strategies to achieve token pruning. These methods can be roughly divided into two paradigms: (1) criterion-based critical token selection \cite{chen2024image,shang2024llava,liu2024multi,huang2024dynamic,xing2024pyramiddrop}, and (2) efficient module design for visual sequence compression \cite{li2024tokenpacker,qian2024spatial,zhang2025llava}.
Despite the progress made by existing approaches in reducing visual token redundancy, they all suffer from severe performance degradation at high pruning rates \footnote{In this work, we define the pruning rate as the proportion of retained visual tokens to the total original visual tokens.} (\textit{e.g.}, pruning rate exceeds 90\%). The phenomenon is intuitively understandable. Since the severely pruned visual tokens preserve minimal visual information, they pose significant challenges to effective multi-modal perception and reasoning.

\setlength{\dashlinedash}{5pt} 
\setlength{\dashlinegap}{3pt}  

\begin{table*}
   \renewcommand\arraystretch{1.0}
    \centering
    \setlength{\tabcolsep}{2pt}   
    \caption{Comparison of our approach with state-of-the-art methods on multiple LVLMs at around 93.5\% pruning rate. `Avg Tokens' denotes the average quantity of visual tokens across all LLM layers. Average tokens of LLaVA-NeXT-7B is computed as the average length of visual sequences on MME benchmark.}
    
    \begin{tabular}{l|c|c|ccccccc|c}
        \cline{1-11}
        Methods & TFLOPs & Avg Tokens & MME & MMBench & POPE & MMVP & SEED$^{\text{img}}$ & GQA & Flickr30k & Average \\
        \cline{1-11}
        LLaVA-1.5-7B \cite{liu2024improved}        & 8.63 (100\%) & 576 (100\%)& 1866.1 & 64.3 & 85.8 & 24.7 & 66.1 & 62.0 & 65.9 & 62.2 \\
        \hdashline 
        + FastV \cite{chen2024image}        & 1.88 (21.8\%) & 70 (12.2\%) & 1337.5 & 54.1 & 42.8 & 10.7 & 49.2 & 48.0 & 32.3 & 40.7 \\
        + VTW \cite{lin2024boosting}          & 1.92 (22.2\%) & 72 (12.5\%)  & 996.2  & 22.3 & 11.7 & 0    & 36.1 & 39.1 & 5.9 & 23.2 \\
        + LLaVolta \cite{chen2024efficient} (test)& 1.88 (21.8\%)& 70 (12.2\%)& 1180.9 & 33.9 & 45.0 & 6.7  & 44.2 & 46.3 & 21.3 & 34.2 \\
        + SparseVLM \cite{zhang2024sparsevlm}    & 1.90 (22.0\%) &  68 (11.8\%)& 1489.6 & 55.8 & 64.1 & 7.3  & 51.6 & 49.5 & 26.2 & 44.0 \\
        + PDrop \cite{xing2024pyramiddrop} (test) & 1.84 (21.3\%) &  66 (11.5\%)& 1428.5 & 52.1 & 55.3 & 11.3 & 49.3 & 49.0 & 33.8 & 43.1 \\
        + ACCM (Ours)           & 1.82 (21.1\%) &  36 (6.3\%)& 1541.9 $\scriptstyle\color{red}{\uparrow52.3}$ & 56.4 $\scriptstyle\color{red}{\uparrow0.6}$& 75.3 $\scriptstyle\color{red}{\uparrow11.2}$ & 16.7 $\scriptstyle\color{red}{\uparrow5.4}$& 54.6 $\scriptstyle\color{red}{\uparrow3.0}$& 52.0 $\scriptstyle\color{red}{\uparrow2.5}$& 36.7 $\scriptstyle\color{red}{\uparrow2.9}$& 49.5 $\scriptstyle\color{red}{\uparrow5.5}$\\
        \cline{1-11}
        LLaVA-1.5-13B \cite{liu2024improved}       & 16.31 (100\%) & 576 (100\%)& 1827.3 & 67.7 & 85.9   & 28.7 & 68.2 & 63.3 & 61.1 & 62.9 \\
        \hdashline 
        + FastV \cite{chen2024image}        & 3.15 (19.3\%) & 63 (10.9\%) & 1506.8 & 59.1 & 56.2 & 10.7 & 54.2 & 51.7 & 39.6 & 46.5 \\
        + VTW \cite{lin2024boosting}          & 3.02 (18.5\%) & 58 (10.1\%)  & 913.9  & 22.8 & 0    & 0.7  & 38.5 & 39.6 & 3.7 & 19.7 \\
        + LLaVolta \cite{chen2024efficient} (test) & 3.15 (19.3\%) & 63 (10.9\%) & 1277.4 & 47.0   & 57.9 & 7.3  & 50.7 & 51.5 & 20.5 & 40.1 \\
        + ACCM (Ours)           & 3.13 (19.2\%) & 36 (6.3\%) & 1519.1 $\scriptstyle\color{red}{\uparrow12.3}$& 60.9 $\scriptstyle\color{red}{\uparrow1.8}$& 73.6 $\scriptstyle\color{red}{\uparrow15.7}$& 17.3 $\scriptstyle\color{red}{\uparrow6.6}$& 55.6 $\scriptstyle\color{red}{\uparrow1.4}$& 53.3 $\scriptstyle\color{red}{\uparrow1.6}$& 36.9 $\scriptstyle\color{green}{\downarrow2.7}$& 50.3 $\scriptstyle\color{red}{\uparrow3.8}$\\
        \cline{1-11}
        LLaVA-NeXT-7B \cite{liu2024llavanext}       & 30.65 (100\%) & 2108 (100\%)& 1846.3 & 67.4 & 86.5 & 38.7 & 70.2 & 64.2 & 44.7 & 62.5 \\
        \hdashline 
        + FastV \cite{chen2024image}        & 5.68 (18.5\%) & 267 (12.7\%)& 1456.0   & 59.4 & 57.4 & 22   & 53.5 & 52.8 & 24.2 & 45.9 \\
        + VTW \cite{lin2024boosting}          & 5.70 (18.6\%) & 264  (12.5\%) & 918.0    & 20.9 & 0.04 & 3.3  & 38.1 & 38.9 & 2.4 & 19.5 \\
        + LLaVolta \cite{chen2024efficient} (test) & 5.68 (18.5\%) & 267 (12.7\%)& 1176.4 & 44.7 & 60.2 & 10.0   & 52.1 & 51.4 & 15.9 & 39.5 \\
        + ACCM (Ours)           & 4.40 (14.4\%) & 144 (6.8\%)& 1620.2 $\scriptstyle\color{red}{\uparrow164.2}$& 59.5 $\scriptstyle\color{red}{\uparrow0.1}$& 85.6 $\scriptstyle\color{red}{\uparrow25.4}$& 25.3 $\scriptstyle\color{red}{\uparrow3.3}$& 61.1 $\scriptstyle\color{red}{\uparrow7.6}$& 56.2 $\scriptstyle\color{red}{\uparrow3.4}$& 31.6 $\scriptstyle\color{red}{\uparrow7.4}$& 53.9 $\scriptstyle\color{red}{\uparrow8}$\\
        \cline{1-11}
    \end{tabular}
    \label{tab:tab-1}
\end{table*}

\subsection{ACCM}   
\label{sec:xxx}
To solve the above problem, we propose an \textbf{A}daptive \textbf{C}ontent \textbf{C}ompensation \textbf{M}ethod (ACCM), which can adaptively supplement the lost information via an image caption. As shown in Figure \ref{fig:pic-2}, ACCM firstly utilizes a lightweight caption model to generate question-related descriptions under the guidance of the user instruction. Then a selector is employed to choose a contextually appropriate caption from multiple candidates.  
Finally, the supplemental caption is provided to the LLM together with the retained visual tokens and the user instruction. In the following, we provide a detailed description of our approach.

\subsubsection{Question-related Caption Generation}   

Given an image $I$, visual token pruning is accomplished after the vision encoder encodes $I$ into a visual sequence $V$. Following \cite{chen2024recoverable}, we utilize [cls] token in $V$ and text embeddings of the user instruction $Q$ to identify critical visual tokens, which are retained after pruning and denoted as $V_{r}$. Other visual tokens are discarded and represented as $V_{l}$.
Notably, alternative token pruning strategies can also be integrated into our approach (\textit{e.g.}, LLaVA-PruMerge \cite{shang2024llava}). 
When the pruning rate is high, $V_{l}$ contains rich visual information, which is helpful for answering user's questions. In contrast to other methods that simply discard them, we convert $V_{l}$ into a brief caption by a lightweight caption model. Following ClipCap \cite{mokady2021clipcap}, we construct the caption model $\mathcal{C}$ consisting of an image encoder, a projector and a language model. For the efficiency purpose, our caption model shares the same encoder (\textit{i.e.}, CLIP image encoder \cite{radford2021learning}) with most of the LVLMs, thus enabling the reuse of visual tokens from LVLMs.

While different questions typically rely on diverse visual cues, a general image caption may not inherently contain these information. To guide $\mathcal{C}$ by user instructions and generate question-related descriptions, $V_{l}$ is combined with $Q$ and processed by the caption model. Considering that generated captions vary in expression and semantic emphasis, with certain formulations probably offering more appropriate content, we generate multiple captions via a specific decoding strategy (\textit{e.g.}, beam search \cite{lowerre1976harpy}) for subsequent processing.
The above process can be formulated as:
\begin{equation}
    M = \mathcal{C}(\operatorname{concat}[Q;V_{l}]),
\end{equation}
where $M = \left\{m_i\right\}_{i=1}^B$ is a set containing $B$ generated captions and $\operatorname{concat}$ represents concatenate operation. Benefiting from the guidance of $Q$, each caption in $M$ describes relevant content accordingly.

\subsubsection{Contextually Appropriate Caption Selection}    

To further get contextually appropriate caption according to the user instruction $Q$, we employ a selector $\mathcal{S}$ to choose from $M$. The selector consists of two components, a text encoder $T$ from LongCLIP \cite{zhang2024long} and a classifier $R$ (instantiated as a transformer block \cite{vaswani2017attention}). 

Given the user instruction $Q$ and caption set $M$, we first concatenate each caption $m_i$ with the $Q$ to form $B$ question-caption pairs and encode them via $T$. Leveraging the summary ability of $T$, the relevance between $m_i$ and $Q$ could be integrated into the [EOS] token embedding of each pair. We formalize the computation as follows: 
\begin{equation}
    z = T(\left\{\operatorname{concat}[Q;m_i] \right\}_{i=1}^B),
\end{equation}
where $z \in \mathbb{R}^{B\times d_z}$ is [EOS] token embeddings from $B$ question-caption pairs.
Then the $R$ is employed to select the most appropriate caption from $z$. The computation can be expressed as:
\begin{equation}
    ind = \operatorname{argmax}(\operatorname{Softmax}(R(z))),
    \label{eq:eq-1}
\end{equation}
where $ind \in \mathbb{R}^{B}$ is a one-hot vector indicates the index of the chosen caption $m_s$ in $M$.
With the help of the selector, we obtain a contextually appropriate caption to provide more proper and precise content.

Finally, retained visual tokens $V_r$, supplemental caption $m_s$ and user instruction $Q$ are provided to LLM for multi-modal comprehension and generation. The above generation process could be formulated as:
\begin{equation}
  p\left(Y \mid V, Q\right)=\prod_{i=1}^L p_{\theta}\left(y_i \mid V_r, m_s, Q, Y, _{<i}\right),
\end{equation}
where $Y$ is the answer with $L$ tokens. Taking advantage of the supplemental caption, ACCM effectively mitigates the severe visual information loss and enhances multi-modal perception and reasoning under high pruning rates.

\subsection{Model Optimization}
\label{sec:optimization}
The optimization objective of our approach is to generate question-related and contextually appropriate captions based on  the corresponding question, which is challenging due to the lack of explicit labels. Meanwhile, Eq.\ref{eq:eq-1} introduces a non-differentiable operation, which is hard to optimize.  Thus we reformulate our training process as a preference optimization task and apply Direct Preference Optimization (DPO) \cite{rafailov2023direct} to optimize our caption model and selector. Moreover, our models are trained in a self-supervised manner without time-consuming labeling.

\subsubsection{Data Construction}
Different from traditional reinforcement learning from human feedback (RLHF) \cite{bai2022training,sun2023aligning}, DPO reparameterizes the policy update using pairwise comparison data ({\it i.e.}, positive samples and negative samples) without training separate reward models. Taking advantage of its convenience, we employ DPO to optimize our models, encouraging them to supplement more relevant and proper information based on the question.

For preparing positive and negative samples, firstly we build a caption set $\textit{A}$ by generating multiple captions via a specific decoding strategy (\textit{e.g.}, beam search \cite{lowerre1976harpy}) conditioned on the discarded tokens and corresponding question. Then we divide $\textit{A}$ into positive caption set $\textit{A}^{+}$ and negative caption set $\textit{A}^{-}$ by comparing their output $O_{+}$ and $O_{-}$ with original output of LVLMs $O_{0}$. To determine the caption that most effectively alleviates the information loss, Kullback-Leibler (KL) divergence \cite{kl} is applied to measure the distance between $O_{+}$, $O_{-}$ and $O_{0}$. 
Finally, we obtain the preference dataset $\textit{D} = \{\textit{A}^{+}, \textit{A}^{-}\}$.

\subsubsection{Optimization Process}
DPO \cite{rafailov2023direct} simplifies the complex policy optimization process of traditional reinforcement learning. It transforms preference learning into a supervised objective by leveraging the implicit reward structure, enabling stable and computationally efficient training. Thus, we apply it to optimize our caption model $\mathcal{C}$ and selector $\mathcal{S}$. Moreover, it has been shown that comparable performance could be achieved without the reference model compared to the vanilla DPO \cite{hong2024reference,meng2024simpo}. Consequently, with the preference dataset $\textit{D} = \{\textit{A}^{+}, \textit{A}^{-}\}$, we define the optimization objectives as follows: 
\begin{equation}
\max _{\mathcal{C}} \mathbb{E}_{\left(x_{c}, \textit{A}^{+}, \textit{A}^{-}\right) \sim \textit{D}} \log \sigma\left(\beta \log \mathcal{C}\left(\textit{A}^{+} \mid x_{c}\right)\right. 
\left.-\beta \log \mathcal{C}\left(\textit{A}^{-} \mid x_{c}\right)\right),
\end{equation}
\begin{equation}
\max _{\mathcal{S}} \mathbb{E}_{\left(x_{s}, \textit{A}^{+}, \textit{A}^{-}\right) \sim \textit{D}} \log \sigma\left(\beta \log \mathcal{S}\left(\textit{A}^{+} \mid x_{s}\right)\right. 
\left.-\beta \log \mathcal{S}\left(\textit{A}^{-} \mid x_{s}\right)\right),
\end{equation}
where $x_{c} = (V_{l}, Q)$ and $x_{s} = (\textit{A}, Q)$ denotes the inputs for $\mathcal{C}$ and $\mathcal{S}$ respectively. The $\beta$ is set to 1 following \cite{meng2024simpo,hong2024reference}. Benefiting from the above optimization objectives, ACCM learns to generate appropriate captions for different questions to mitigate information loss under high pruning rates. More importantly, our approach keeps the weights of LVLMs frozen and only optimize the caption model $\mathcal{C}$ and selector $\mathcal{S}$, which greatly reduces the training overheads.

\begin{table*}
   \renewcommand\arraystretch{1.0}
    \centering
    \setlength{\tabcolsep}{2pt}   
    \caption{Comparison of our approach with state-of-the-art methods on multiple LVLMs at around 97\% pruning rate. `Avg Tokens' denotes the average quantity of visual tokens across all LLM layers. Average tokens of LLaVA-NeXT-7B is computed as the average length of visual sequences on MME benchmark.}
    \begin{tabular}{l|c|c|ccccccc|c}
        \cline{1-11}
        Methods & TFLOPs & Avg Tokens & MME & MMBench & POPE & MMVP & SEED$^{\text{img}}$ & GQA & Flickr30k & Average \\
        \cline{1-11}
        LLaVA-1.5-7B \cite{liu2024improved}          & 8.63 (100\%) & 576 (100\%)& 1866.1 & 64.3 & 85.8 & 24.7 & 66.1 & 62.0 & 65.9 & 62.2\\
        \hdashline 
        + FastV \cite{chen2024image}        & 1.66 (19.2\%) & 53 (9.2\%) & 1120.5 & 39.4 & 19.5 & 4.0  & 40.8 & 41.6 & 15.1 & 28.6 \\
        + VTW \cite{lin2024boosting}          & 1.68 (19.5\%) & 54 (9.4\%)  & 956.8  & 21.4 & 0.084 & 1.3  & 35.6 & 38.8 & 4.6 & 23.7 \\
        + LLaVolta \cite{chen2024efficient} (test)& 1.66 (19.2\%)& 53 (9.2\%)& 1049   & 28.1 & 40.2 & 3.3  & 40.3 & 42.6 & 13.9 & 29.4 \\
        + SparseVLM \cite{zhang2024sparsevlm}    & 1.66 (19.2\%) &  49 (8.5\%)& 1291.5 & 47.9 & 43.9 & 6.0  & 47.6 & 46.0 & 15.5 & 36.1 \\
        + PDrop \cite{xing2024pyramiddrop} (test) & 1.69 (19.6\%) &  55 (9.5\%)& 1293.4 & 48.6 & 42.8 & 9.3  & 47.5 & 46.1 & 25.0 & 37.9 \\
        + ACCM (Ours)           & 1.58 (18.3\%) &  18 (3.1\%)& 1333.8 $\scriptstyle\color{red}{\uparrow40.4}$& 52.4 $\scriptstyle\color{red}{\uparrow3.8}$& 70 $\scriptstyle\color{red}{\uparrow26.1}$  & 18.7 $\scriptstyle\color{red}{\uparrow9.4}$& 50.8 $\scriptstyle\color{red}{\uparrow3.2}$& 49.4 $\scriptstyle\color{red}{\uparrow3.3}$& 30.8 $\scriptstyle\color{red}{\uparrow5.8}$& 45.7 $\scriptstyle\color{red}{\uparrow7.8}$\\
        \cline{1-11}
        LLaVA-1.5-13B \cite{liu2024improved}         & 16.31 (100\%) & 576 (100\%)& 1827.3 & 67.7 & 85.9   & 28.7 & 68.2 & 63.3 & 61.1 & 62.9\\
        \hdashline 
        + FastV \cite{chen2024image}        & 2.72 (16.7\%) & 46 (8.0\%) & 1305.5 & 46.2 & 33.2 & 4.7 & 47.0 & 46.3 & 19.9 & 34.8\\
        + VTW \cite{lin2024boosting}          & 2.65 (16.2\%) & 43 (7.5\%)  & 898.4  & 22.2 & 0.03 & 0.7  & 38.1 & 39.5 & 3.6 & 19.5 \\
        + LLaVolta \cite{chen2024efficient} (test) & 2.72 (16.7\%) & 46 (8.0\%) & 1124.7 & 38.7 & 49.9 & 5.3  & 46.1 & 47.4 & 14.2 & 34.5 \\
        + ACCM (Ours)           & 2.67 (16.4\%) & 18 (3.1\%) & 1374.8 $\scriptstyle\color{red}{\uparrow69.3}$& 56.1 $\scriptstyle\color{red}{\uparrow9.9}$& 68.4 $\scriptstyle\color{red}{\uparrow18.5}$ & 14.7 $\scriptstyle\color{red}{\uparrow9.4}$& 51.2 $\scriptstyle\color{red}{\uparrow4.2}$& 50.5 $\scriptstyle\color{red}{\uparrow3.1}$& 32.5 $\scriptstyle\color{red}{\uparrow12.6}$& 46.1 $\scriptstyle\color{red}{\uparrow11.3}$\\
        \cline{1-11}
        LLaVA-NeXT-7B \cite{liu2024llavanext}         & 30.65 (100\%) & 2108 (100\%)& 1846.3 & 67.4 & 86.5 & 38.7 & 70.2 & 64.2 & 44.7 & 62.5 \\
        \hdashline 
        + FastV \cite{chen2024image}        & 4.80 (15.7\%) & 199 (9.4\%) & 1233.3 & 45.3 & 25.7 & 16.0   & 45.6 & 45.1 & 9.0 & 33.0 \\
        + VTW \cite{lin2024boosting}          & 4.81 (15.7\%) & 198 (9.4\%)  & 914.3  & 20.4 & 0.07 & 2.0    & 37.5 & 38.7 & 3.6 & 19.3 \\
        + LLaVolta \cite{chen2024efficient} (test) & 4.80 (15.7\%) & 199 (9.4\%) & 1015.9 & 33.9 & 38.9 & 5.3  & 45.8 & 46.3 & 9.3 & 30.8 \\
        + ACCM (Ours)           & 3.47 (11.3\%) & 72 (3.4\%) & 1484.4 $\scriptstyle\color{red}{\uparrow251.1}$& 54.9 $\scriptstyle\color{red}{\uparrow9.6}$& 83.1 $\scriptstyle\color{red}{\uparrow44.2}$& 23.3 $\scriptstyle\color{red}{\uparrow7.3}$& 56.0 $\scriptstyle\color{red}{\uparrow10.2}$ & 52.7 $\scriptstyle\color{red}{\uparrow6.4}$& 16.0 $\scriptstyle\color{red}{\uparrow6.7}$& 48.4 $\scriptstyle\color{red}{\uparrow15.4}$\\
        \cline{1-11}
    \end{tabular}
    \label{tab:tab-2}
\end{table*}

\section{Experiments}
\subsection{Benchmarks and Metrics}
We conduct extensive experiments across a range of multi-modal tasks to validate the effectiveness of our approach, including  visual question answering, visual reasoning, hallucination evaluation and image caption. The benchmarks we employed include MME \cite{mme}, MMBench \cite{liu2024mmbench}, MMVP \cite{tong2024eyes}, POPE \cite{li2023evaluating}, SEED-Bench \cite{li2023seed}, GQA \cite{hudson2019gqa} and Flickr30k \cite{young2014image}. 
For Flickr30k, we report the CIDEr score \cite{vedantam2015cider} as the evaluation metric. 
For other benchmarks, the official metrics are used.

\subsection{Implementation Details}
We apply our approach to three prevalent LVLMs, including LLaVA-1.5-7B \cite{liu2024improved}, LLaVA-1.5-13B \cite{liu2024improved} and LLaVA-Next-7B \cite{liu2024llavanext}. The weights of LVLMs are fixed in all experiments.  
We construct our caption model following ClipCap \cite{mokady2021clipcap}, with 143.5M parameters.
For the selector, we utilize text encoder of LongCLIP \cite{zhang2024long} to encode question-caption pairs for effectively processing long context. A four-layer transformer is employed to instantiate the classifier. 
During self-supervised training, only the language model of caption model and the classifier of selector are learnable. LLaVA-1.5-7B is employed to accomplish token pruning and generate output logits by default.  We randomly sample 40k data from LLaVA-665k dataset \cite{liu2024improved} to construct the preference dataset. 
During inference, the selector chooses from three captions generated via beam search \cite{lowerre1976harpy}. 
Following FastV \cite{chen2024image} and PyramidDrop \cite{xing2024pyramiddrop}, we report the TFLOPs and average quantity of visual tokens. All experiments are conducted on a single NVIDIA A100 40G GPU.

\subsection{Main Results}
In Table \ref{tab:tab-1}, we conduct extensive experiments on multiple prevalent LVLMs (\textit{i.e.}, LLaVA-1.5-7B \cite{liu2024improved}, LLaVA-1.5-13B \cite{liu2024improved} and LLaVA-NeXT-7B \cite{liu2024llavanext}) and compare our approach with existing methods under high pruning rates (\textit{i.e.}, 93.5\% and 97\%). The comparison methods include FastV \cite{chen2024image}, VTW \cite{lin2024boosting}, LLaVolta \cite{chen2024efficient}, SparseVLM \cite{zhang2024sparsevlm} and PyramidDrop \cite{xing2024pyramiddrop}, which accomplish token pruning without modifying the weights of LVLMs.   
As shown in Table \ref{tab:tab-1}, our approach obviously surpasses other methods across seven benchmarks at 93.5\% pruning rate on LLaVA-1.5-7B \cite{liu2024improved}. For instance, ACCM achieves 49.5\% on average, outperforming SparseVLM \cite{zhang2024sparsevlm} by an absolute increase of 5.5\% with lower FLOPs. Compared to FastV \cite{chen2024image}, ACCM obtains a more obvious performance gain of 8.8\%. The significant advantages demonstrate that our method could efficiently supplement the lost visual information and alleviate performance degradation on various multi-modal tasks. When applied to LLaVA-1.5-13B \cite{liu2024improved} and LLaVA-NeXT-7B \cite{liu2024llavanext}, ACCM maintains its leading performance, exhibiting the robustness of our approach. Specifically, on LLaVA-NeXT-7B \cite{liu2024llavanext}, the average result of our approach exceeds FastV \cite{chen2024image} by 17.4\%, while reducing FLOPs by 22.5\%. Especially on POPE \cite{li2023evaluating} benchmark, ACCM achieves an F1 score of 85.6\%, with only a 0.9\% performance drop compared to unpruned result, while reducing FLOPs from 30.65T to 4.4T. 

In Table \ref{tab:tab-2}, we further evaluate our approach under a extreme pruning rate (around 97\%). As shown in Table \ref{tab:tab-2}, the advantages of our approach become more distinct. ACCM obtains an average absolute improvement of 7.8\% on LLaVA-1.5-7B \cite{liu2024improved} compared to PDrop (PyramidDrop) \cite{xing2024pyramiddrop}. On LLaVA-1.5-13B \cite{liu2024improved} and LLaVA-NeXT-7B \cite{liu2024llavanext}, ACCM achieves absolute performance gains of 11 and 15 percentage points respectively. The consistent superiority across different high pruning rates validates the effectiveness of our method in mitigating visual information loss. The efficiency analysis of our approach and comparison methods can be found in Supplementary Material.

\begin{table}
   \renewcommand\arraystretch{1.0}
    \centering
    \caption{Ablation study on the key components. `Baseline' refers to the setting where we apply token pruning without our approach. `+ caption' represents that the supplemental caption is directly generated without the guidance of corresponding question and selector. }
    
    \begin{tabular}{c|l|cc}
        \cline{1-4}
        & Method  & MME & POPE \\ 
        \cline{1-4}
        1 & LLaVA-1.5-7B \cite{liu2024improved}  & 1866.1 & 85.8 \\
        \hdashline 
        2 & Baseline     & 1463.3 & 66.7 \\
        3 & + caption    & 1486.7 & 70.2 \\
        4 & + question guidance    & 1516.7 & 73.1 \\
        5 & + selector   & 1541.9 & 75.3 \\
                
        \cline{1-4}
    \end{tabular}
    \label{tab:tab-3}
\end{table}

\subsection{Ablation Study}
\label{ablation}
In Tabel \ref{tab:tab-3}, we ablate the key components of ACCM. The experiments are conducted on MME \cite{mme} and POPE \cite{li2023evaluating} with LLaVA-1.5-7B \cite{liu2024improved}. The pruning rate is set to 93.5\%. In the first row of Table \ref{tab:tab-3}, we report the original results of LLaVA-1.5-7B. As shown in row 2 of Table \ref{tab:tab-3}, applying token pruning under high pruning rate causes severe performance degradation, with 402.8 and 19.1\% performance drops on MME and POPE respectively. As shown in row 3, a general image caption could supplement the lost information and alleviate the degradation to some extent, with 23.4 and 3.5\% performance gains on MME and POPE compared to baseline. In row 4, we guide the caption model by corresponding questions to generate relevant descriptions and the performance improves compared to supplementing a general caption (+30.0, +2.9\% on MME and POPE correspondingly).  Based on generated question-related captions, we further employ a selector to choose a contextually appropriate description from multiple candidates. As shown in row 5, the selector further mitigates the degradation. The improvements over baseline and the variant using a general caption (\textit{i.e.}, row 3) validate the effectiveness of our approach in generating relevant descriptions and supplementing appropriate information to mitigate performance degradation under high pruning rates.

\begin{table}
   \renewcommand\arraystretch{1.0}
    \centering
    \caption{The performance of our approach with diverse caption models. In the evaluation, LLaVA-1.5-7B is employed and 36 visual tokens are retained after pruning (at 93.5\% pruning rate).}
    \begin{tabular}{l|ccc}
        \cline{1-4}
        Model                  & MME    & MMB  & GQA  \\
        \cline{1-4}
        UniversalCap \cite{cornia2021universal}   & 1526.3 & 56.8 & 51.5   \\
        ClipCap \cite{mokady2021clipcap}        & 1541.9 & 56.4 & 52.0   \\
                
        \cline{1-4}
    \end{tabular}
    \label{tab:tab-6}
\end{table}

\begin{figure}
  \centering
    \includegraphics[width=0.45\textwidth]{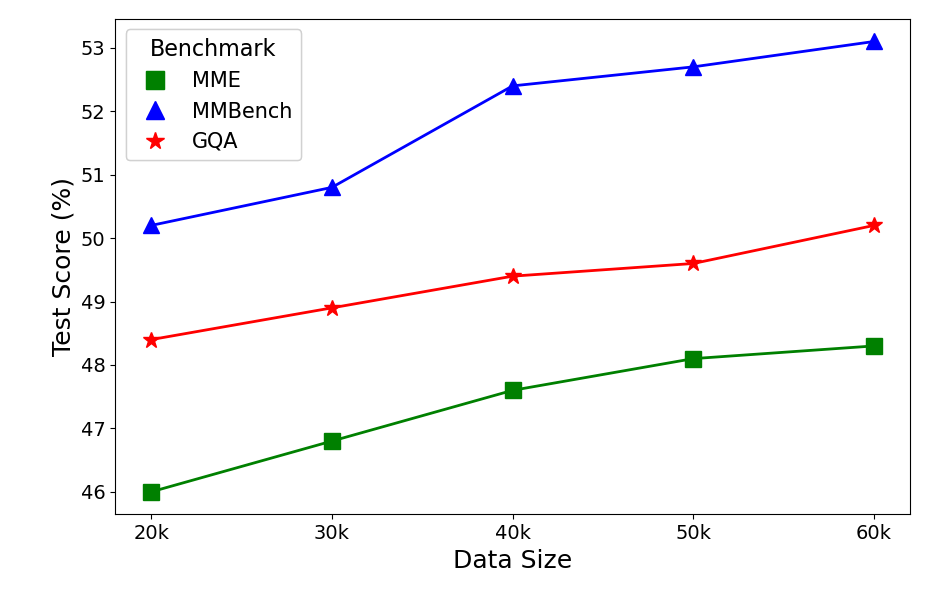}
   
  \caption{The impact of training data size in self-supervised training. LLaVA-1.5-7B is employed and pruning rate is set to 97\% in evaluation. We normalize the test score of MME for illustrative purposes.}
  \label{fig:pic-6}
\end{figure}

\begin{table}
   \renewcommand\arraystretch{1.0}
    \centering
    \setlength{\tabcolsep}{1pt}   
    \caption{Cross-model transferability of our approach. We employ different LVLMs to train our modules and directly apply them to another LVLM in evaluation. The pruning rate is set to 97\% in evaluation.}
    \begin{tabular}{cccc|cccc}
        \cline{1-7}
        \multicolumn{2}{c}{Self-supervised Training} & \multicolumn{2}{c}{Evaluation}  & \multicolumn{1}{|c}{\multirow{2}*{MME}} & \multirow{2}*{MMB} & \multirow{2}*{GQA} \\
        LLaVA$^{\text{1.5}}$ & LLaVA$^{\text{NeXT}}$ & LLaVA$^{\text{1.5}}$ & LLaVA$^{\text{NeXT}}$ &  &  &   \\
        \cline{1-7}
        \checkmark  &   & \checkmark  &   & 1333.8 & 52.4 & 49.4  \\
          & \checkmark  & \checkmark  &   & 1310.9 & 52.0 & 48.3  \\
        \cline{1-7}
        \checkmark  &   &   & \checkmark  & 1484.4 & 54.9 & 52.7  \\
          & \checkmark  &   & \checkmark  & 1451.2 & 53.8 & 52.6  \\
        
        \cline{1-7}
    \end{tabular}
    \label{tab:tab-4}
\end{table}

\subsection{Alternative Settings}
In this section we systematically investigate three key factors: (1) robustness to caption model variations, (2) the volume of training data in self-supervised learning, and (3) cross-model generalization of our approach.

\subsubsection{Caption Model Substitution}
To examine the influence of diverse caption models in our approach, we substitute a different caption model, UniversalCap \cite{cornia2021universal} for ClipCap \cite{mokady2021clipcap} in Table \ref{tab:tab-6}. UniversalCap could effectively utilize noisy image-caption pairs without compromising the descriptive style of human-annotated datasets. Unlike ClipCap (143.5M params), UniversalCap employs a encoder-decoder architecture with fewer parameters (We employ the Tiny version with 105M params). As shown in Table \ref{tab:tab-6}, ClipCap achieves an average score of 54.5\% on three benchmarks (\textit{i.e.}, MME \cite{mme}, MMBench \cite{liu2024mmbench} and GQA \cite{hudson2019gqa}) and UniversalCap obtains 54.3\%, which are at the same level. The comparable results achieved by different caption models demonstrate that our approach is flexible and robust, without relying on any specific model. 

\subsubsection{Training Data Size}   
To explore the impact of the training data size in self-supervised learning, we scale up the data volume from 20k to 60k in Figure \ref{fig:pic-6}. In the experiments, LLaVA-1.5-7B is employed during training and evaluation. As shown in Figure \ref{fig:pic-6}, scaling up the training data brings performance gains across three benchmarks consistently. For instance, when the training data size increases from 20k to 60k, our approach achieves +2.3\%, +2.9\% and +1.8\% gains on MME \cite{mme}, MMB \cite{liu2024mmbench} and GQA \cite{hudson2019gqa}, respectively. With the training data enlarged, the caption model becomes better at generating question-related descriptions and the selector could choose more proper captions to supplement the lost information.

\subsubsection{Transferability across LVLMs}   
To assess the transferability of our approach, we train our modules with one LVLM in self-supervised training and directly apply them to another LVLM without additional tuning. As shown in Table \ref{tab:tab-4}, we employ LLaVA-1.5-7B \cite{liu2024improved} and LLaVA-NeXT-7B \cite{liu2024llavanext} in training respectively. It can be observed from the second row that when transferred from LLaVA-NeXT-7B to LLaVA-1.5-7B, our approach achieves comparable results (49.8\% \textit{vs.} 49.0\%) to the variant without cross-model transfer (\textit{i.e.}, the first row of Table \ref{tab:tab-4}). Likewise, when transferred from LLaVA-1.5-7B to LLaVA-NeXT-7B, the performance is also on par with the non-transfer variant (the fourth row of Table \ref{tab:tab-4}). 
The consistently strong performance across different LVLMs demonstrates the cross-model generalization ability of our approach. As a result, our proposed caption model and selector could be transferred to diverse LVLMs in a plug-and-play manner.

\subsection{Performance across Diverse Pruning Rates}
\label{performance degradation}
\begin{figure}
  \centering
    \includegraphics[width=0.48\textwidth]{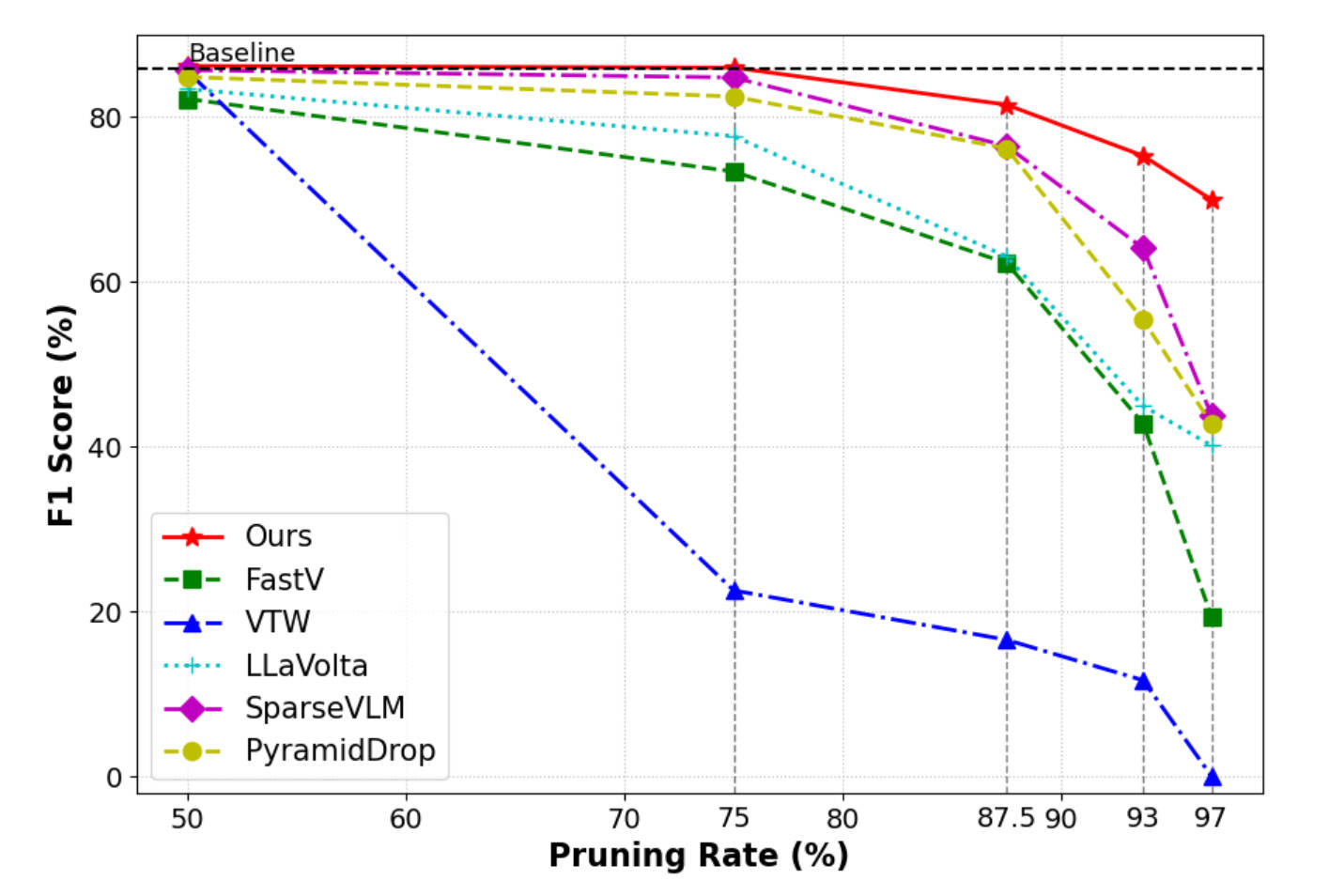}
   
  \caption{Performance comparison of our approach and existing methods across varying pruning rates. The experiment is conducted upon LLaVA-1.5-7B and evaluated on POPE.}

  \label{fig:pic-1_1}
\end{figure}

In Figure \ref{fig:pic-1_1}, we demonstrate the performance of our approach and existing methods across diverse pruning rates. As shown in Figure \ref{fig:pic-1_1}, when the pruning rate is below 75\%, most comparison methods maintain the performance. While their performance begins to drop noticeably once the pruning rate exceeds 75\%. After the pruning rate goes beyond 90\%, all comparison methods experience a sharp degradation in performance. For instance, FastV \cite{chen2024image} suffers a drop of more than 30\%. In contrast, our approach not only maintains strong performance under lower pruning rates, but also effectively mitigates performance deterioration at high pruning rates compared to other methods (outperforming SparseVLM \cite{zhang2024sparsevlm} by 26.1\% at 97\% pruning rate). The consistent superiority across varying pruning rates validates the robustness of our approach.

\begin{figure}
  \centering
    \includegraphics[width=0.49\textwidth]{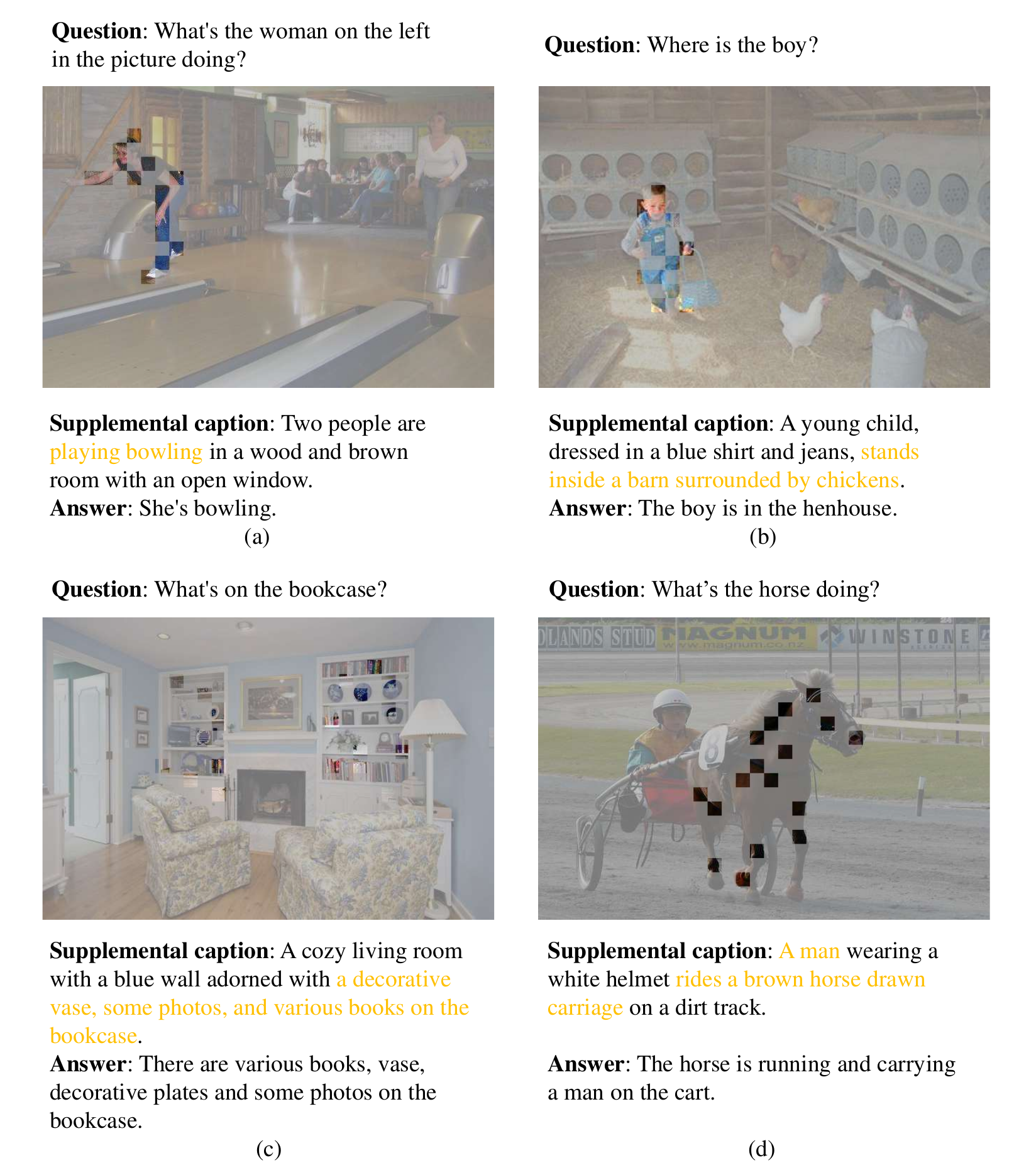}     
   
  \caption{Visualization of token pruning results and supplemental captions. The answer is generated by  our approach with LLaVA-1.5-7B. The relevant content to the corresponding question in captions is highlighted in yellow.}

  \label{fig:pic-vis}
\end{figure}

\subsection{Qualitative Analysis}
To visually demonstrate the effectiveness of our approach in supplementing the lost information, we visualize token pruning results and supplemental captions of our approach on some samples. As shown in Figure \ref{fig:pic-vis} (a), at 97\% pruning rate, LVLMs entirely focus on the woman and lose sight of bowling balls behind her, which makes it difficult to infer the woman's activity. While our approach could effectively recover the lost information (\textit{i.e.}, playing bowling) by generating a question-related caption. In Figure \ref{fig:pic-vis} (c), there are various items on the bookcase and LVLMs miss some of them with few visual tokens retained. Our method could generate multiple relevant descriptions and select the most proper one (\textit{i.e.}, the caption which covers more items on the bookcase) to support multi-modal perception and reasoning.

\section{Conclusion}
In the paper, we introduce \textbf{A}daptive \textbf{C}ontent \textbf{C}ompensation \textbf{M}ethod (ACCM), aiming to alleviate the visual information loss for efficient LVLMs. When encountering high pruning rates, current token pruning methods suffer from serious performance degradation due to severe visual information loss. To tackle the problem, we propose to adaptively supplement the lost information by an image caption. 
Our approach consists of two key components: a lightweight caption model and a selector. Guided by the user instruction, the caption model could generate question-related descriptions firstly. Then the selector identifies the most contextually appropriate caption from multiple candidates. Finally our models are joint optimized by DPO. More importantly, our approach can be optimized in a self-supervised manner, without any human or automated labeling.
Extensive experiments across multiple benchmarks and LVLMs validate the superiority of ACCM, surpassing other methods significantly with lower FLOPs. 
To the best of our knowledge, ACCM is the first work that leverages image captions to recover the visual information lost during token pruning and we hope it can provide a new thought for the community.

\begin{acks}
This work is supported by the National Natural Science Foundation of China (No. U23B2013).
\end{acks}

\bibliographystyle{ACM-Reference-Format}
\balance
\bibliography{sample-sigconf}










\end{document}